\documentclass[IJPHM, 2019, 0]{ijPHMSociety}

\usepackage{graphicx}
\usepackage{amsmath}
\usepackage[noend]{algpseudocode}
\usepackage{upquote}

\begin{document}

% Paper Title
\title{Deep Detector Health Management under Adversarial Campaigns}

% Authors List
\author{%			
	Javier Echauz\authorNumber{1}, Keith Kenemer\authorNumber{2}, Sarfaraz Hussein\authorNumber{3}, Jay Dhaliwal\authorNumber{4}, Saurabh Shintre\authorNumber{5},\\ Slawomir Grzonkowski\authorNumber{6}, and Andrew Gardner\authorNumber{7}
}

% Author Affiliations
\address{% This is a tabular environment so each affiliation needs to be separated by "\\" or "\tabularnewline"
	\affiliation{{1-3,7}}{Symantec Corporation, Atlanta, GA, 30328, USA}{ %add emails
		{\email{\{Javier\_Echauz}},
		{\email{Keith\_Kenemer}},
		{\email{Sarfaraz\_Hussein}},
		{\email{Andrew\_Gardner\}@symantec.com}}
		} % emails input
	\tabularnewline % skip one row for next affiliation	
	\affiliation{{4,5}}{Symantec Corporation, Mountain View, CA, 94043, USA}{ % add emails
		{\email{\{Jasjeet\_Dhaliwal}}, {\email{Saurab\_Shintre\}@symantec.com}}
		} % emails input
	\tabularnewline % skip one row for next affiliation	
	\affiliation{6}{Symantec Corporation, Dublin, Ireland}{ % add emails
		{\email{Slawomir\_Grzonkowski@symantec.com}}
		} % emails input
}

% Create the title
\maketitle

% PHM Society Distribution License Information, provide first author's name "FirstName LastName"
\phmLicenseFootnote{Javier Echauz}

% Abstract
\begin{abstract}
Machine learning models are vulnerable to adversarial inputs that induce seemingly unjustifiable errors. As automated classifiers are increasingly used in industrial control systems and machinery, these adversarial errors could grow to be a serious problem. Despite numerous studies over the past few years, the field of adversarial ML is still considered alchemy, with no practical unbroken defenses demonstrated to date, leaving PHM practitioners with few meaningful ways of addressing the problem. We introduce \textit{turbidity detection} as a practical superset of the adversarial input detection problem, coping with adversarial \textit{campaigns} rather than statistically invisible one-offs. This perspective is coupled with ROC-theoretic design guidance that prescribes an inexpensive domain adaptation layer at the output of a deep learning model during an attack campaign. The result aims to approximate the Bayes optimal mitigation that ameliorates the detection model's degraded health. A proactively reactive type of prognostics is achieved via Monte Carlo simulation of various adversarial campaign scenarios, by sampling from the model's own turbidity distribution to quickly deploy the correct mitigation during a real-world campaign.
\end{abstract}

\section{Introduction}

A machine learning application often begins with a dataset of examples and the task is to find a classification model that will turn inputs into class-label predictions, while preserving some sense of minimum expected error. The learning problem is often \textit{unrealizable}, so no perfect model exists that will have 0 generalization error \cite{shalev2014understanding}. But less obviously, it is often possible to deterministically find input examples that \textit{force} the model to misclassify \cite{szegedy2014intriguing}. Machine learning (ML) models can be subjected to adversarially crafted small perturbations that purposely induce these errors, and they can seem unjustified or surprising to a human observer (e.g., a digital image of a school bus mistaken for a bird). As automated ML-based classifiers pervade across applications in transportation, medicine, finance, and cybersecurity, adversarial errors could grow to be a very serious problem. The danger is particularly acute in industrial control systems (ICS), industrial Internet of Things (IIoT), automation equipment, and factory robotics, where malfunctions can be life-threatening (e.g., steel mill furnace explosions, power grid crashes, etc.). Unfortunately, ICS attacks are on the rise, with increased vectors for malicious party access to critical infrastructure \cite{icscert17}. Detection of attacks to cyberphysical systems \cite{yan2018cyberattack}, and particularly as it relates to adversarial ML, is a growing area of concern that has been underserved in PHM literature.

Despite vigorous study over the past few years (see review in \cite{gilmer2018motivating}), the field of adversarial ML is considered by researchers to be at a nascent stage \cite{evanstalk}, with no practical unbroken defenses demonstrated to date (attacks succeed with $p>.25$) \cite{carlini2017adversarial}, and still talks of an ``arms race'' between attackers and defenders \cite{goodfellow2018making}. This leaves PHM practitioners with few meaningful ways of addressing the problem. We have identified a fundamental flaw in the current interpretation of adversarial defenses, and offer an alternative practical reformulation of the problem that copes with population-level \textit{campaign mitigation} as opposed to individual input, case-by-case protection.

The defense side of adversarial ML has tried to answer a blend of two questions: (a) How to \textit{robustify} a model (make it harder for an attacker to fool)? This has led to adversarial training, defensive distillation, feature squeezing, architecture modification, and minimax optimization \cite{madry2017towards}; and (b) What can be \textit{measured} about adversarial inputs that is different from regular ones? This has led to input validators and adversarial detectors \cite{goodfellow2018making}. Generative adversarial networks (GANs) can synthesize adversarial examples which can then be used to retrain the classifier, however, this only helps insofar as it gets a classifier closer to Bayes optimality (it can also make things worse). By omissions in the current discourse, these methods have created the illusion that we could one day prebake a solution at training time that will protect a model against \textit{one-or-few-off} adversarial inputs at deployment time. Our work suggests that the latter goal comes at a disproportionate price in expected error. Intuitively, if there was a way to accurately detect error-inducing inputs at runtime, then that same detector would have been used to augment or improve the training to begin with.

In the following sections, we will introduce \textit{turbidity detection} as a different, ROC-centric way of thinking about adversarial example detection that fixes current widespread misinterpretations and leads to a practical mitigation. Our theory yields 3 previously unreported results: (i) unqualified use of an adversarial detector inverts ROC (harms); (ii) adversarial campaign pinches down ROC (harms); and (iii) conditions exist where the ROC can be repaired to at least a gracefully degraded state during the campaign. We propose a methodology for putting that into practice and show experimental results using image (digit recognition) and IIoT security (malware detection) data.

%Path relative to the .tex file containing the \includegraphics command
\graphicspath{ {images/} }
\begin{figure*}[h]
\centering
\includegraphics[width=7in]{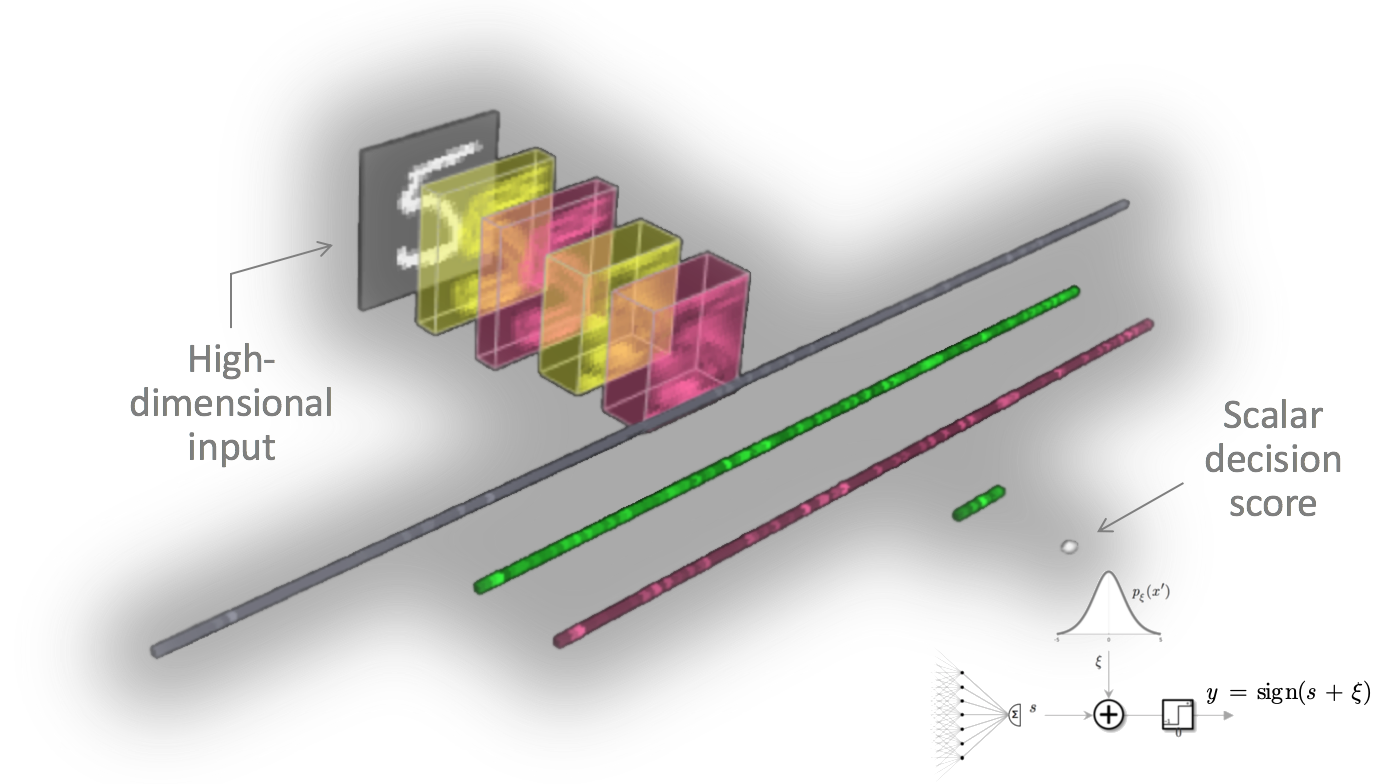}
\caption{Deep neural network from input to scalar decision score s with discrete-choice true label $y$.}
\label{fig:nnet}
\end{figure*}

\section{Turbidity Detection Theory}

Our first aim is to show that the unqualified use of an adversarial detector has deleterious effect on ROC. To that end, we will start in a seemingly restrictive setting: 1-dimensional input, uniform distribution over --10 to 10, binary output from binomial discrete-choice theory with logistic noise, equiprobable classes, and Bayes decision rule. However, our main results (ROC inversion, pinching, and repair) will not critically depend on these specific choices, retaining clarity of illustration without loss of generality.

Instead of asking where adversarial examples are ``hiding'' in high-dimensional input space, we focus on the scalar decision score output axis (preactivation/logit or post-activation/ pseudo-probability), where model-processed samples have to end up anyway, and where any decision confusion actually occurs. Figure 1 shows a deep neural network taking an input array through convolutional and nonlinear activation layers, then dense layers reducing the output to a scalar decision score $s$ (here logit). Consider a data-generating process (DGP) such that ground-truth bipolar labels $y \in \{-1, +1\}$ come from adding the score to a $Logistic(0,1)$ symmetric noise $\xi$ (whose scale and bias control class separability and class imbalance respectively), and taking sign:

\begin{equation}
    y = {\rm{sign}}(s + \xi )\;.
\end{equation}

The symmetry of the noise about its 0 mean implies our DGP emits equiprobable labels: $P(0)=P(1)=0.5$. In order to output a monotonic and correctly calibrated posterior probability $P(1|s)$, what the last-layer activation of the deep net ``wants to be'' is the CDF of the discrete-choice noise:

\begin{equation}
\begin{array}{l}
P(y = 1|s) = \int\limits_0^\infty  {{p_\xi }(s' - s)} ds'\\
 = \int\limits_{ - s}^\infty  {{p_\xi }(s')} ds' = 1 - \int\limits_{ - \infty }^{ - s} {{p_\xi }(s')} ds'\\
 = 1 - CD{F_\xi }( - s) = CD{F_\xi }(s)\,.
\end{array}
\end{equation}

The logistic-distributed noise has logistic sigmoid CDF, agreeing with an output neuron (here with $\mu=0, c=1$):

\[\xi  \sim Logistic(\mu ,c)\;\;  \Rightarrow CD{F_\xi }(s) = \frac{1}{{1 + {e^{ - \left( {\tfrac{{s - \mu }}{c}} \right)}}}} = \sigma \left( {\tfrac{{s - \mu }}{c}} \right).\]

The Bayes-optimal decision rule (one yielding least probability of misclassification in our DGP) corresponds to the homogeneous halfspace (here semiaxis) obtained by thresholding the above posterior probability at 0.5, or directly thresholding the preactivation score:

\setcounter{equation}{3}
\begin{equation} \label{eq:4}
\hat y = {\mathop{\rm sign}\nolimits} (s)\:.
\end{equation}

\begin{figure*}[h]
\centering
\includegraphics[width=7in]{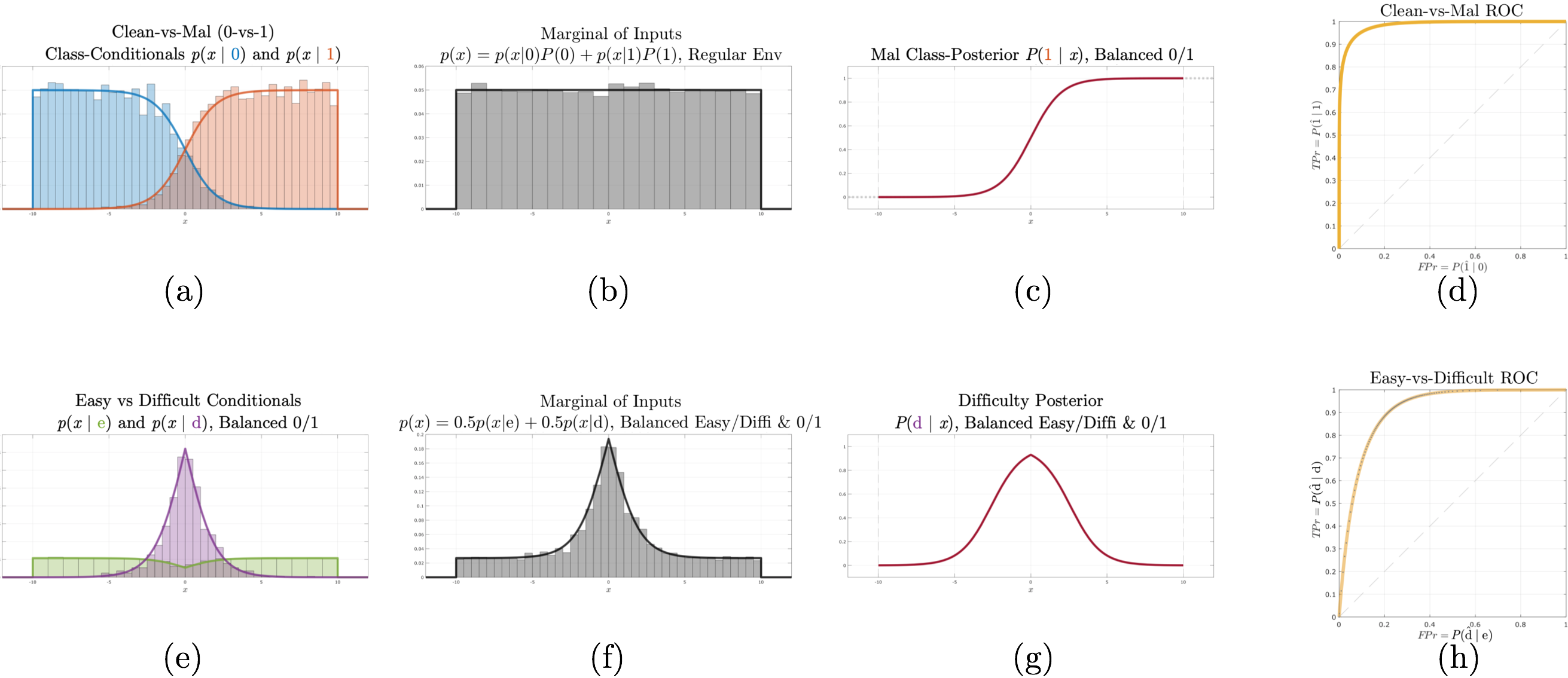}
\caption{Top: Class-conditionals, marginal, posterior, and ROC of 0-vs-1 detector in regular environment. Bottom: Class-conditionals, marginal, posterior, and ROC of corresponding e-vs-d detector.}
\label{fig:0-vs-1&e-vs-d}
\end{figure*}

Now we derive the ROC for this ideal detector in its regular environment. From Bayes theorem, the 0 (``clean'')-vs-1 (``mal'') class-conditionals of the score are

\begin{equation} \label{eq:5}
\begin{array}{l}
p(s|0) = {F_\xi }( - s)\,{\cal U}(s; - 10,10)/0.5\:,\\
p(s|1) = {F_\xi }(s)\,{\cal U}(s; - 10,10)/0.5\:,
\end{array}
\end{equation}

(see Figure 2(a)) where $F(\cdot)$ denotes CDF from now on, ${\cal U}(s;$\newline--10,10) = $[–10\le s \le 10]/20$ is the uniform PDF, and that last Iverson bracket [$\cdot$] means indicator function: valued 1 when the event $s$-within-the-interval is true and 0 otherwise. In our 1D mathematical reference figures, the score is directly equal to the input: $s=x$ (while in higher dimensions, it will be an inner product where coordinates can be explanatory variables, features, previous neural layers, etc.).

The marginal of scores is the uniform PDF (Figure 2(b)), while the malicious class-posterior is (Figure 2(c)):

\begin{equation} \label{eq:6}
P(1|s) = \frac{{p(s|1) \cdot {\textstyle{\frac{1} {2}}}}}{{p(s|0) \cdot {\textstyle{\frac{1} {2}}} + p(s|1) \cdot {\textstyle{\frac{1} {2}}}}} = {F_\xi }(s)\:.
\end{equation}

Finally, monotonicity of the class-1 posterior allows us to obtain the ROC from a single sweep on the s-axis, yielding the parametric curve (Figure 2(d)):

\begin{equation} \label{eq:7}
(FPr,\:TPr) = \left( {1 - {F_0}(s),\:\:1 - {F_1}(s)} \right)\:,
\end{equation}

where

\begin{equation} \label{eq:8}
{F_0}(s) = \int\limits_{ - \infty }^s {p(s'|0)ds'} \:,\quad {F_1}(s) = \int\limits_{ - \infty }^s {p(s'|1)ds'} \:.
\end{equation}

\subsection{Clarity and Turbidity Distributions}

Unless classes are 100\% separable in a generalization preserving way relative to the DGP (input features, label noise, and their statistical relation), every model, including the Bayes-optimal one, experiences difficulty whenever it makes the wrong class prediction. We say that samples that confuse the model, i.e., FPs and FNs, are \textbf{turbid} from the model's point of view, whereas all the other correctly-classified TNs and TPs are \textbf{clear}. We can think of every model that tackles the original 0-vs-1 problem as having an inherent dual problem: separating clear-vs-turbid (denoted \textbf{e}-vs-\textbf{d} as mnemonic for ``\textbf{e}asy''-vs-``\textbf{d}ifficult''), for which a different detector can be built. Since the model's confusion depends on its output threshold, by default we peg the associated turbidity detection concept to the maximum balanced-accuracy/Youden index threshold in the original detector

\begin{equation} \label{eq:8b}
\theta = \mathop {\arg \max }\limits_{\theta  \in [0,1]} \{TPr(\theta) - FPr(\theta)\} \:,
\end{equation}

i.e., the ROC operating point closest to upper-left corner.

Next we present the clarity and turbidity distributions for a DGP where there is 50-50\% proportion of clear vs turbid samples (something that we will characterize as a \textit{toxic} environment compared to the regular one where mistakes should be rare), and 50-50\% proportion of clean vs mal within each. Obtain each conditional as a mixture of the truncated class-0 plus the truncated class-1 PDFs. For example, the left half of turbidity $p(s |\rm{d})$ consists of the left tail of $p(s | 1)$ (= FNs) normalized by the area under it up to 0 (= $F_1(0)$), while the right half has the right tail of $p(s | 0)$ (= FPs) normalized by the area under it from 0 onward (= $1-F_0(0) = F_1(0) =.0693$). The mixture of these 2 densities then gives the inflexed arch shape (purple in Figure 2(e)), similar to a truncated Laplace distribution:

\begin{equation} \label{eq:9}
\begin{array}{l}
p(s|{\rm{e}}) = \frac{{{\textstyle{\frac{1}{2}}}p(s|0)[s < 0] + {\textstyle{\frac{1}{2}}}p(s|1)[s \ge 0]}}{{{F_0}(0)}}\:,\\\\
p(s|{\rm{d}}) = \frac{{{\textstyle{\frac{1}{2}}}p(s|0)[s \ge 0] + {\textstyle{\frac{1}{2}}}p(s|1)[s < 0]}}{{{F_1}(0)}}\:.
\end{array}
\end{equation}

Obtain the marginal of scores from the equiprobable mixture of the clear and turbid distributions (or from total probability theorem; Figure 2(f)), and the turbidity class-posterior as (Figure 2(g)):

\begin{equation} \label{eq:10}
P({\rm{d}}|s) = \frac{{p(s|{\rm{d}}) \cdot \frac{1}{2}}}{{p(s|{\rm{e}}) \cdot \frac{1}{2} + p(s|{\rm{d}}) \cdot \frac{1}{2}}}\:.
\end{equation}

Note that this symmetric reverse-ogee arch is nonmonotonic. This implies that the theoretical ROC curve can no longer be obtained simply by sweeping a single threshold over the $s$ domain; doing so would result in a suboptimal improper curve (under diagonal chance line). The most general method is to sweep a descending threshold on the vertical axis of the class-posterior, nonlinearly solve/root-find all critical $s$ values where posterior intersects the threshold, then calculate area under class-conditionals over $s$ regions so as to obtain the pair $(FPr,TPr)$. In effect, the ROC curve computation becomes multibranched, with number of connected segments dependent on number of intersections encountered during the sweep. A general multibranched algorithm is given in Appendix A1. Figure 2(h) shows the exact ROC, using either the multibranched algorithm just described or an alternative monotonic version afforded by symmetry in this case. Luckily, when data scientists compute an empirical ROC (i.e., from a data sample), they automatically obtain a Monte Carlo estimate, so theoretical complications like the nonmonotonicity above are never encountered. However, the scores should be presented as the possibly nonmonotonic posteriors instead of as preactivations.

\subsection{Relation to Adversarial Detection}

The widely accepted {\it oracle} definition of adversarial examples \cite{evanstalk} states that: (i) they are created with intent to deceive, (ii) they start from a seed example of say class A, correctly seen as class A by the model, and (iii) after perturbation they still behave like class A according to the oracle/ground-truth, yet they are now incorrectly seen as class B by the model. However, the goal of “adversarial example detection” (accurately determining at runtime whether an input is adversarial) has been widely misconstrued, leading to overfitting and/or invalidly-dichotomized detectors. If we insist we can detect a particular set of adversarial samples, then that same detector is bound to fail on a freshly created one operating in a regular environment. It will work if operated in a toxic environment, but then for a whole different reason as we'll see below.

A typical adversarial detection experiment starts from a dataset of regular samples, takes each instance in the dataset as a seed to which a transformation (e.g., from CleverHans library \cite{papernot2016technical}) is applied in order to create an adversarial counterpart, and then sees if the ``regular-vs-adversarial'' examples are discernible in some way (e.g., by showing differences in distributions or by building adversarial detectors and measuring their above-chance discrimination). By definition, \textit{all adversarial examples are turbid}. Further, they can exist with ``high model confidence'' (with $P(0|x)$ or $P(1|x)$ near 1). But exactly the same is true of natural, unforced errors. All regular FPs and FNs are turbid, and while most are associated with low confidence ($P(0|x)=P(1|x)=0.5$) near model's decision boundary, high-confidence ones also arise. They happen as predicted even by the 1D DGP, just less frequently, consistent with the tapered-but-still-nonzero tails of the turbidity distribution. Thus, regular and adversarial samples can share the same domain.

We don't believe human intent is a distinguishing feature that can be measured either---a view hinted in \cite{carlini2017adversarial}---anymore than telling if the person who made the samples was left-handed from looking at the numerical input coordinates. So what is it in the standard adversarial detection experiment that is being detected? \underline{Answer}: the observed difference between regular and adversarial conditions stems from the fact that \textit{all} adversarial samples are turbid/difficult by definition, whereas in the regular environment turbid samples are \textit{rare}. Turbid samples tend to concentrate while regular $\approx$ clear tend to spread, thus second moments separate. A scientific animation illustrating this point can be seen in \cite{javier10ml}.

These issues can be fixed by moving from a fortuitous ``regular-vs-adversarial'' dichotomy to the principled ``clear-vs-turbid,'' and by not blurring the line between detector and its intended deployment environment \cite{gilmer2018motivating}. Dropping intent and seed-of-origin out of the adversarial character makes the problem realistic and applicable to campaign mitigation.

\subsection{ROC Inversion}

We now show that a realistic adversarial (i.e., e-vs-d) detector cannot actionably help 0-vs-1 decision-making in a regular environment as it leads to \textbf{ROC inversion}. In the spirit of reductio ad absurdum, let the theoretical 1D adversarial detector $A(x)$ in Figures 2(g,h), Eq.~(\ref{eq:10}) augment the probabilistic 0-vs-1 detector $D(x)$ in Figures 2(c,d), Eq.~(\ref{eq:6}). Given any input $x'$ at test time, if $A(x')$ is accurately declaring that $x'$ is adversarial then we would want to contradict the decision from $D(x')$. From the point of view of $A(x)$, the Bayes trigger to declare adversariness is $A(x)>0.5$, which is equivalent to checking if input magnitude is within a critical cutoff: $|x|<2.597$ (also equal to the crossover points in Figure 2(e)). The augmented detector becomes

\begin{equation}
\hat y = {\mathop{\rm sign}\nolimits} (x) {\mathop{\rm sign}\nolimits} (\left| x \right| - {\rm{2.597)}}\,,
\end{equation}

and the augmented-system posterior probability is:

\[{P_{{\rm{aug}}}}(1|x) = {F_\xi }(x)[\left| x \right| > {\rm{2.597}}] + {F_\xi }( - x)[\left| x \right| \le {\rm{2.597}}]\:.\]

(Any function that reverses the $D(x)$ decision within that interval will work.) Figure 3(a) shows this nonmonotonic posterior. Figure 3(b) shows the exact ROC using multibranched algorithm. The original accuracy of $F_0(0)P(0)+(1-F_1(0))P(1)$ = 0.93 goes down to 0.79. The original detector was already optimal for its intended regular environment, and overriding its decisions only makes it worse. Thus, {\it protection against one-off adversarial examples is a misguided design goal}.

\begin{figure}[h]
\centering
\includegraphics[width=3.38in]{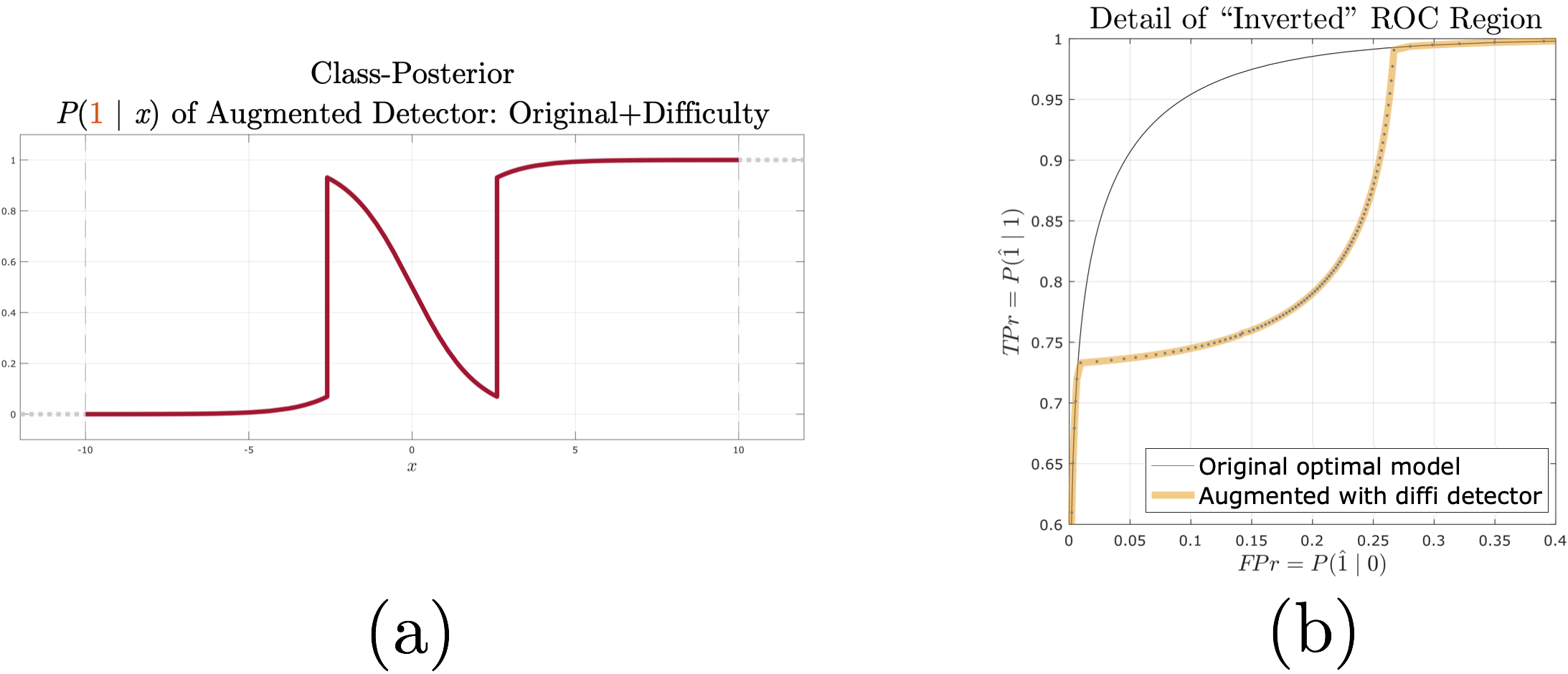}
\caption{Posterior and ROC of the augmented original + turbidity detector leading to ROC inversion.}
\label{fig:ROC_inversion}
\end{figure}

\subsection{ROC Pinch-Down}

We now show that operating the original 0-vs-1 detector in a toxic environment leads to an \textbf{ROC pinch-down}. Under adversarial campaign, class-conditionals can display abnormal concentrations around the original decision score threshold, a single crossover (as in Figure 2(a)) can become multiple, class-posterior can turn nonmonotonic, and errors become frequent, making model performance plummet. Continuing with the balanced proportions of 50-50\% e-vs-d and 50\%-50\% 0-vs-1 of Section 2.1, the class-conditional likelihoods that the original detector now has to confront are (Figure 4(a)):

\setcounter{equation}{13}
\begin{equation}
\begin{array}{l}
{p_{{\rm{toxic}}}}(x|0) = {p_{\rm{e}}}(x)[x \ge 0] + {p_{\rm{d}}}(x)[x < 0]\:,\\
{p_{{\rm{toxic}}}}(x|1) = {p_{\rm{e}}}(x)[x < 0] + {p_{\rm{d}}}(x)[x \ge 0]\:,
\end{array}\end{equation}

where $p_{\rm e}$ and $p_{\rm d}$ are the conditionals in Eq.~{\ref{eq:9}}.

\begin{figure}[h]
\centering
\includegraphics[width=3.38in]{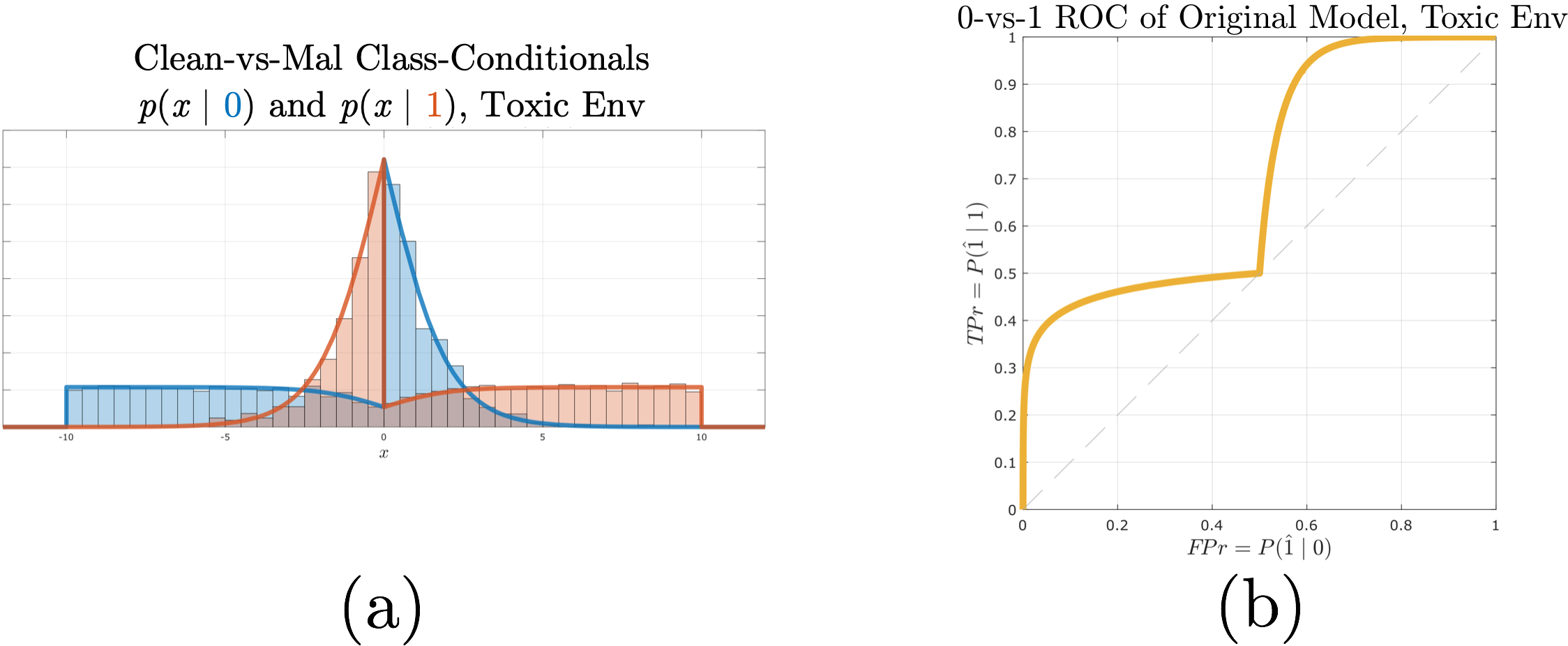}
\caption{Class-conditionals in toxic environment leading to ROC pinch-down.}
\label{fig:ROC_pinchdown}
\end{figure}

The marginal of inputs is identical to Figure 2(f), just composed differently from the average of the above toxic conditionals. The posterior (not shown) is the same one in Figure 2(c) since the model remained naively unchanged during this toxic campaign. The exact ROC (Figure 4(b)) can be obtained here from the monotonic sweep form in Eq.~\ref{eq:7}, except with $F_{0\rm{,toxic}}$ and $F_{1\rm{,toxic}}$. So while augmenting the detector during normal operation was harmful, ignoring the problem during abnormal operation is also potentially worse. Thus, {\it protection against adversarial campaigns (not one-offs) is needed}.

\subsubsection{Asymmetric Toxic Environments}

At the 1:1 ratio of e-vs-d samples under both 0 and 1 classes, the characteristic ROC ``seagull'' (Figure 4(b)) has curve pinned at the chance line (50\% accuracy). However, at other ratios of turbidity proportions under each class, conditionals become asymmetric and the pinch point moves somewhere else. Class-0 e:d ratio controls the horizontal axis (FPr), while class-1 e:d ratio independently controls the vertical axis ($TPr = 1-FNr$). This means that if adversarial campaign actors could not only add samples but also subtract from the environment seen by the model, they would be able to place the pinch-down point anywhere on the ROC plane! But they would have to be oracles themselves, for example, to force the model to be always wrong in the future would pin operating point at the bottom-right corner (something we can't do ourselves with imperfect knowledge). Figure 5 shows a toxic formulation where class-0 samples are regular (i.e., no adversarial FPs), with their natural $F_0(0):F_1(0)$ proportion of clear to turbid, whereas class-1 samples have an unnatural 37.5\%-62.5\% proportion.

\begin{figure}[h]
\centering
\includegraphics[width=3.38in]{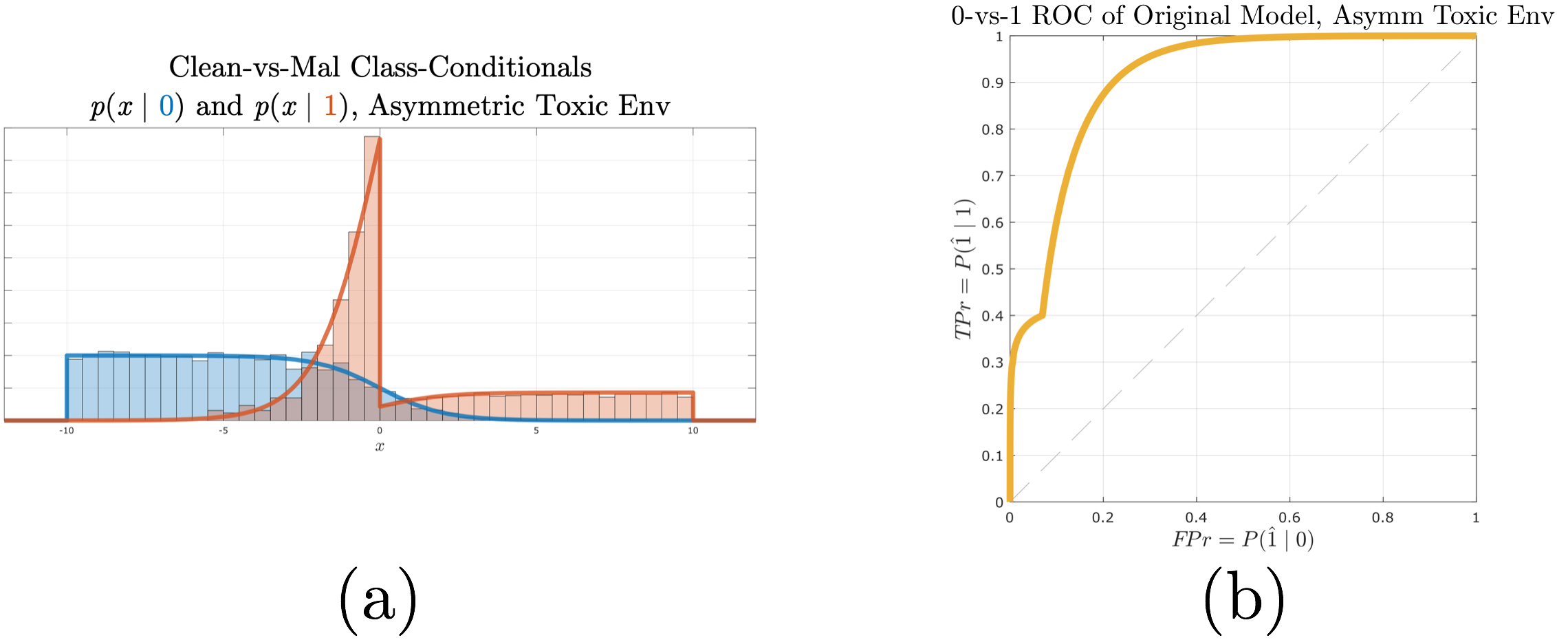}
\caption{Class-conditionals in asymmetric toxic environment, moving ROC pinch-down somewhere else.}
\label{fig:ROC_pinchdown_asymm}
\end{figure}

\subsection{Mitigation/Repair of the Degraded ROC}

In its intended regular environment, the original model can adapt to changes in maliciousness imbalance (0-vs-1 prevalences) by simply sliding its operating point along the intact, class prevalence-agnostic ROC curve. However, in the adversarially toxic environment it is no longer enough to simply adjust a threshold to match the environment; a fundamentally different detection problem must be solved. In order to ``unpinch'' the ROC to the best available shape given the adversarial campaign, we should obey the new posterior:

\begin{equation} \label{eq:15}
{P_{{\rm{toxic}}}}(1|x) = \frac{{{p_{{\rm{toxic}}}}(x|1)P(1)}}{{{p_{{\rm{toxic}}}}(x|0)P(0) + {p_{{\rm{toxic}}}}(x|1)P(1)}}\:.
\end{equation}

This will typically be nonmonotonic (Figure 6(a)). The exactly repaired ROC is obtained from the multibranched algorithm as shown in Figure 6(b).

\begin{figure}[h]
\centering
\includegraphics[width=3.38in]{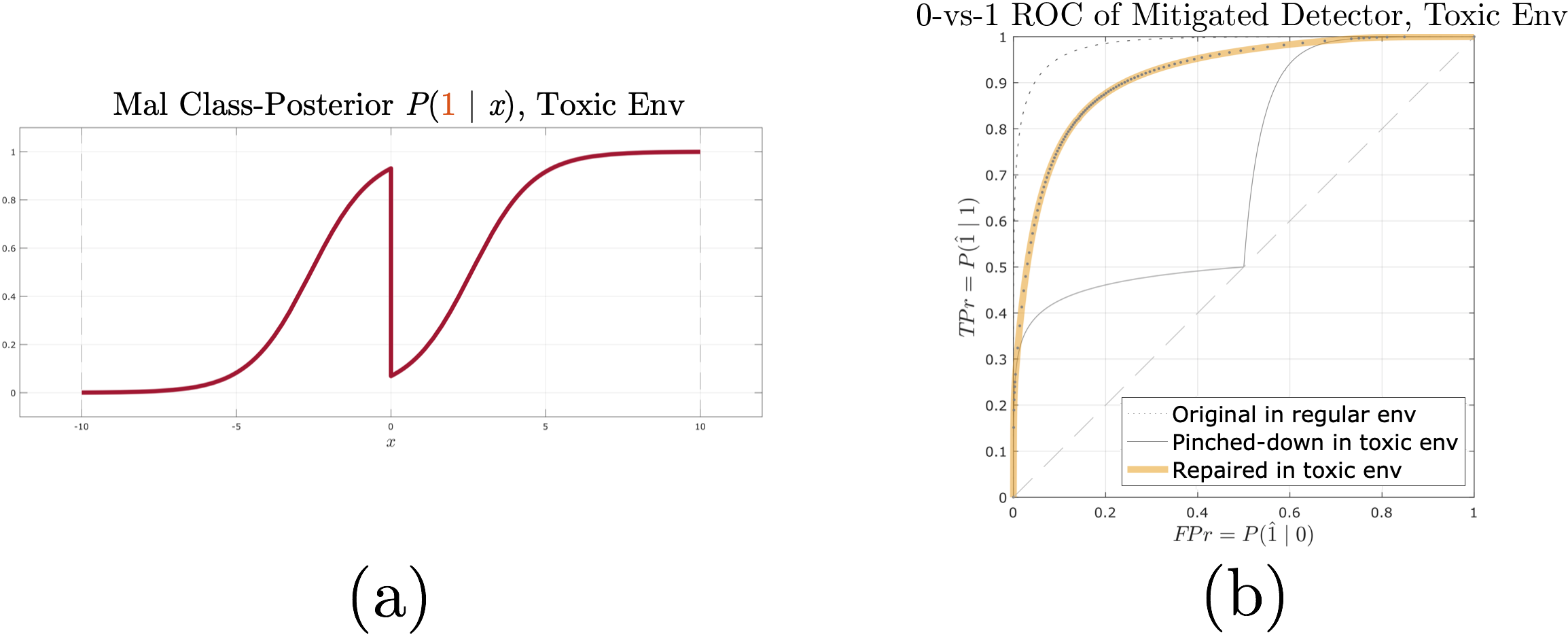}
\caption{New posterior and mitigated/repaired ROC.}
\label{fig:ROC_repair}
\end{figure}

The new optimal maximum-a-posteriori Bayes classifier implements \textit{decision reversals} relative to the original one. Reversals occur only within the decision score intervals where the new heights of 0-vs-1 conditional likelihoods have swapped their dominance, due to the new concentration of turbid/difficult samples in the environment. Thus, {\it the mitigated detector is gracefully (rather than catastrophically) degraded, restoring acceptable error rates and adaptability to maliciousness imbalance}.

\subsubsection{When Repair Isn't Possible}

In some cases it isn't really possible to ``unpinch'' the ROC because the curve morphs into a seamless one with no dent (as if in Figure 5(b) the pinched point fused into the left branch), e.g., with ratio of e-vs-d samples still at 1:1 but with malicious class prevalences falling outside of the interval $F_1(0) < P(1) < F_0(0)$. The curve is still depressed compared to the original regular one due to 50\% of samples being turbid, and only detection threshold remains as a potential adjustment.

We have also uncovered \textbf{adversarial covariate shift} as another condition where ROC repair isn't possible. This would make score class-conditionals and marginal more turbid while keeping the posterior intact. For example,

\[\begin{array}{l}
{p_{{\rm{toxic}}}}(x|0) = \left( {1 - {F_\xi }(s)} \right) \cdot \left( {0.5p(x|{\rm{e}}) + 0.5p(x|{\rm{d}})} \right)/0.5\:,\\
{p_{{\rm{toxic}}}}(x|{\rm{1}}) = {F_\xi }(s) \cdot \left( {0.5p(x|{\rm{e}}) + 0.5p(x|{\rm{d}})} \right)/0.5\:.
\end{array}\]

The reader can verify that the posterior $P(1|x)$ is exactly recovered as $F_\xi(x)$, the CDF of the label noise (also true for other imbalanced 0-vs-1 priors). However, it seems unrealistic that an adversary could shape the conditionals in this fashion as it would require omnipotent control of the environment beyond merely adding adversarial samples to the regular one seen by the model.

\begin{figure*}[h]
\centering
\includegraphics[width=7in]{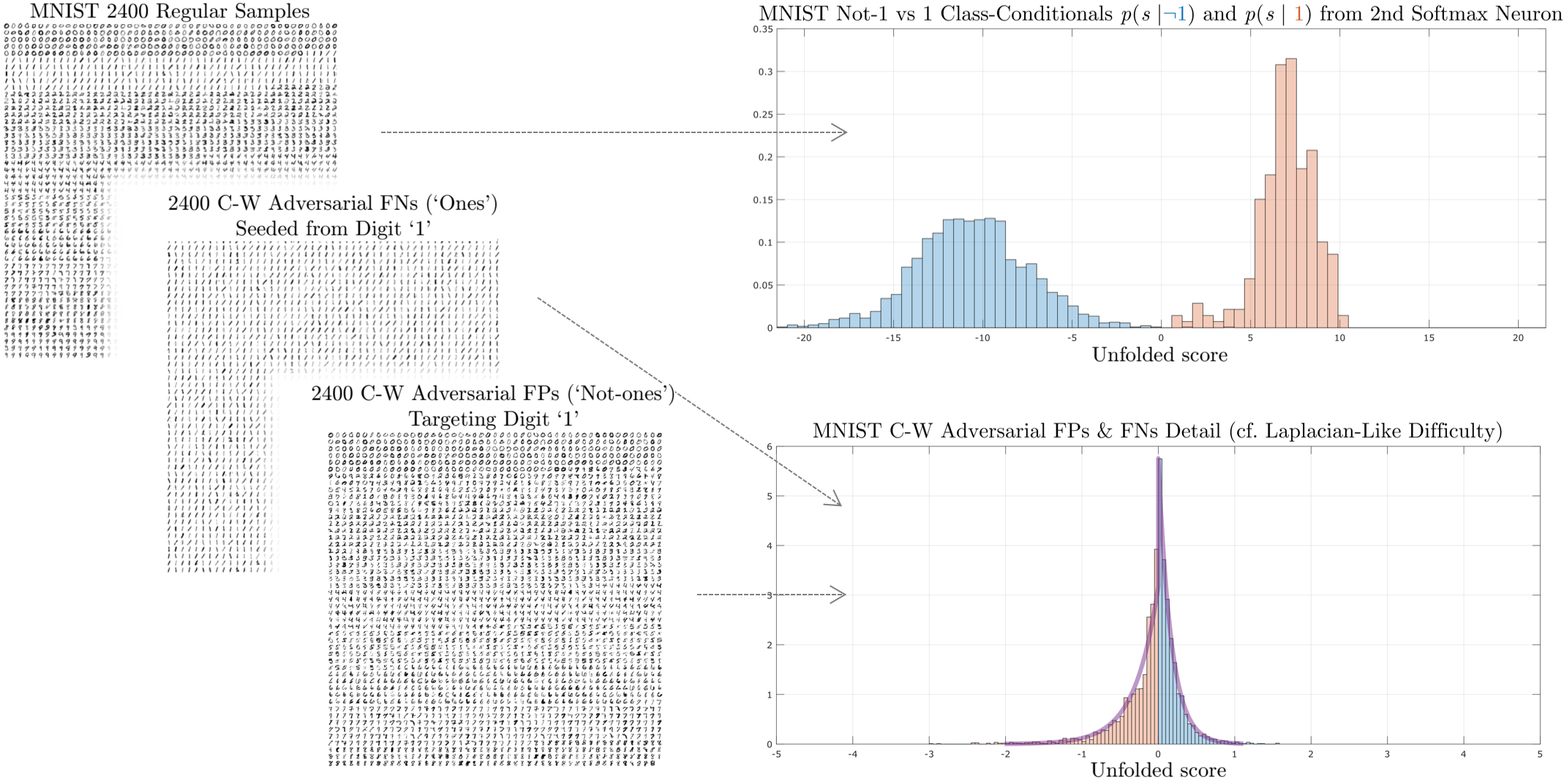}
\caption{Regular not-1 vs 1 conditional histograms and balanced turbidity histogram from adversarial FPs \& FNs.}
\label{fig:MNIST}
\end{figure*}

\subsection{Generalization to Suboptimal Models and Higher Dimensions}

We have systematically charted an atlas documenting how the above 1D reference theory is impacted when the model is suboptimal instead of Bayes-optimal (via mistuned bias and/or misaligned weights), and in higher-dimensional input space, where the decision score $s$ is taken to be the preactivation ${\textbf{\textit{x}}^ \top }\textbf{\textit{w}}$, i.e., the (possibly augmented) dot product at the output layer of a probabilistic binary classifier. Due to space restrictions, we only mention that all ROC inversion, pinch-down, and mitigation repair results remain qualitatively identical. The score is still 1D; the only difference is that with independent components of $\textbf{\textit{x}}$, all distributions become windowed/tapered, and ROCs get ``dumber''/shallower from the CLT centrality effect of marginal $p(\textbf{\textit{x}})$, which makes samples appear close to decision boundary more frequently. Injecting correlation structure in components of $\textbf{\textit{x}}$ also weakens separability, but the main ROC results hold. Further, nothing above prevents decision scores from being computed by \textit{non}-neural network models. Thus, the described campaign effects and mitigation apply to any decision-making component that exposes its scores to attackers, including ensembles of decision trees widely prevalent in industrial settings.

\section{Preemptive Domain Adaptation}

The theoretical results in the previous section can be put into PHM practice by monitoring estimates of the decision score class-conditional distributions in order to declare if and when an adversarial campaign is in effect, repair the degraded ROC during campaign, restore the original model after campaign subsides, and improve readiness for future attacks via simulation. Assume a well-trained classifier has been deployed in its originally-intended threat environment where errors are rare (e.g., $>95$\% hit at $<0.1$\% FP rates). A health management methodology can track 0-vs-1 conditional score histograms (and optionally error rates), from which class-conditionals curves are kernel-density estimated (KDE) as smooth functions. This still requires ground-truth label estimates; in cybersecurity they are obtained after some lag ranging from sub-seconds (with access to cloud-based reputation, etc.) to days (offline endpoints with sporadic live updates, air-gapped IIoT devices, etc.). There is also a way to detect without ground truth, by introspectively looking at whether too many decision scores are falling in a low-confidence interval, but this is bypassed if attackers actively inject only high-confidence samples.

Under adversarial campaign, class-conditionals may develop multimodality, with multiple crossover points that misfit the original decision rules, making model performance plummet for some period of time. This condition can be declared from observation of empirical ROC pinch-down or abnormally large error rates. (In the low-confidence campaign case, the model can track its own decision scores falling in an interval near decision threshold at higher-than-historical rates, suggesting adversarial manipulation since it would be rare to see that in the regular environment.) At that time, an equiprobable class-posterior function implementing Eq. (15) is transmitted to the endpoint model to be used as a post-transformation layer of the original decision scores. The output of this function can then be thresholded to obtain a desired mitigated ROC operating point. In effect, this is an inexpensive statistical domain adaptation that reverses decisions when it makes sense to do so. The same logic can be applied in reverse to restore the original model when campaign has subsided.

The methodology above is still 100\% reactive defense. Our investigation suggests that a 100\% proactive defense (where the model is hardened at training time against all future one-off adversarial samples in the regular environment) is mathematically impossible. Thus, prognostics in the usual sense of predicting remaining life until failure, to do something about it before it occurs, is outside the scope of our work. However, we introduce a {\it proactively reactive} compromise. It precomputes the optimal response to each of several plausible adversarial attack scenarios, via Monte Carlo simulation drawing from the model's own turbidity distribution, and stores that information as a look-up table to quickly deploy the correct mitigation during a real-world campaign. A final health maintenance modality is to put the decision modification layer into effect continuously/ prophylactically without waiting to detect that a campaign has begun (in which case the last layer calculation automatically yields simply an identity function). This way, as machine operating conditions change even gradually, the method is already there to mitigate possibly harmful effects while any persistent shifting is investigated.

\section{Experiments with Real-World Data}

This section verifies the main ROC inversion, pinching, and repair results using real-world data with corresponding attacks against a deep neural network classifier in two application areas: digit recognition and IIoT malware detection.

\subsection{Digit Recognition}

The standard MNIST benchmark dataset was used, containing 60,000 grayscale $28\times28$px images of handwritten digits. The deep convolutional neural network trained in \cite{dhaliwal2018gradient}, whose first layers are visualized in Figure 1, achieved over 99\% accuracy on a holdout split of the data. A stratified random sample of 2400 images was taken to equally represent all digits. We adversarially generated 2400 FPs and 2400 FNs using the Carlini-Wagner algorithm \cite{carlini2017towards}. The 10-class problem was dichotomized into classes `not-1' vs `1' by unfolding the preactivation decision scores as

\setcounter{equation}{16}
\begin{equation}
s = \frac{1}{2}({s_1} - \max \{ {s_0},{s_2},{s_3},{s_4},{s_5},{s_6},{s_7},{s_8},{s_9}\} \,,\end{equation}

where $s_1$ is the preactivation score at the neuron for class `1' (2nd indicator in softmax layer). That leaves 2160 regular instances of class `not-1' and 240 of `1'---a class-prior imbalance of 9:1. Figure 7 shows the pdf-normalized regular conditional histograms (top) and the turbidity distribution from the adversarial FNs and FPs (bottom; additionally color-coded by not-1 vs 1 classes). As predicted by the theory, the latter distribution has a Laplace-like inflex concentration around the score decision-crossing point (cf. purple in Figure 2(e)). For clarity, it is shown with balanced not-1s vs 1s within the turbid condition; the regular environment would have 9 times more FPs than FNs while toxic ones can be manipulated. In this potentially overfit ``99\%'' accuracy case, we cannot display an empirical turbidity distribution with only the natural FPs and FNs because there were only 2 and 0 cases, respectively.

\begin{figure}[h]
\centering
\includegraphics[width=3.38in]{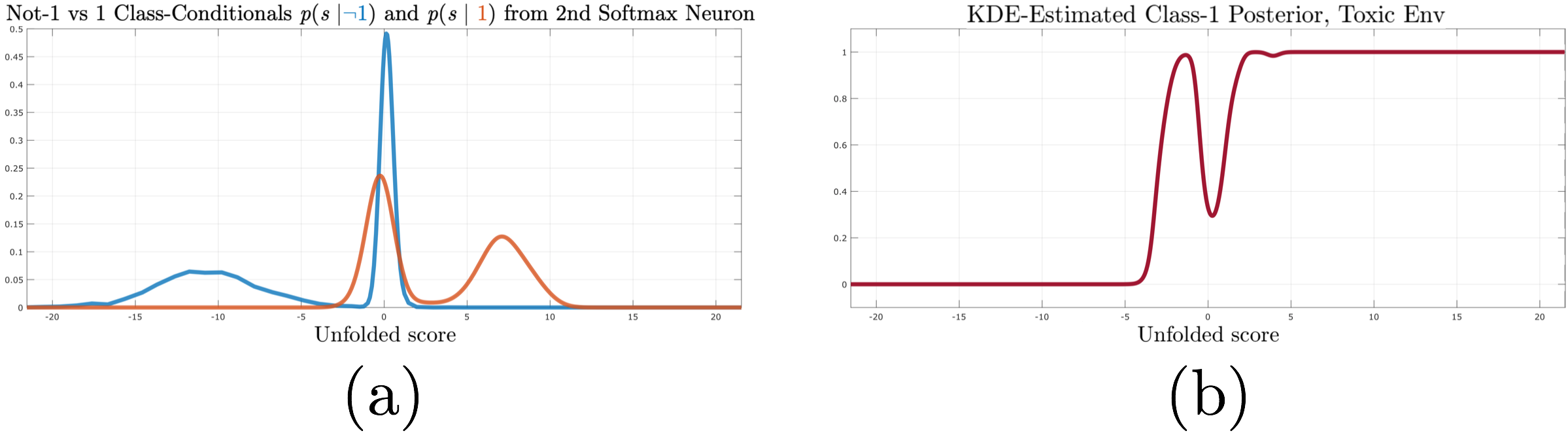}
\caption{KDE-smoothed class conditionals and corresponding equiprobable posterior for use in repair.}
\label{fig:KDE_smoothed}
\end{figure}

\begin{figure}[h]
\centering
\includegraphics[width=3.38in]{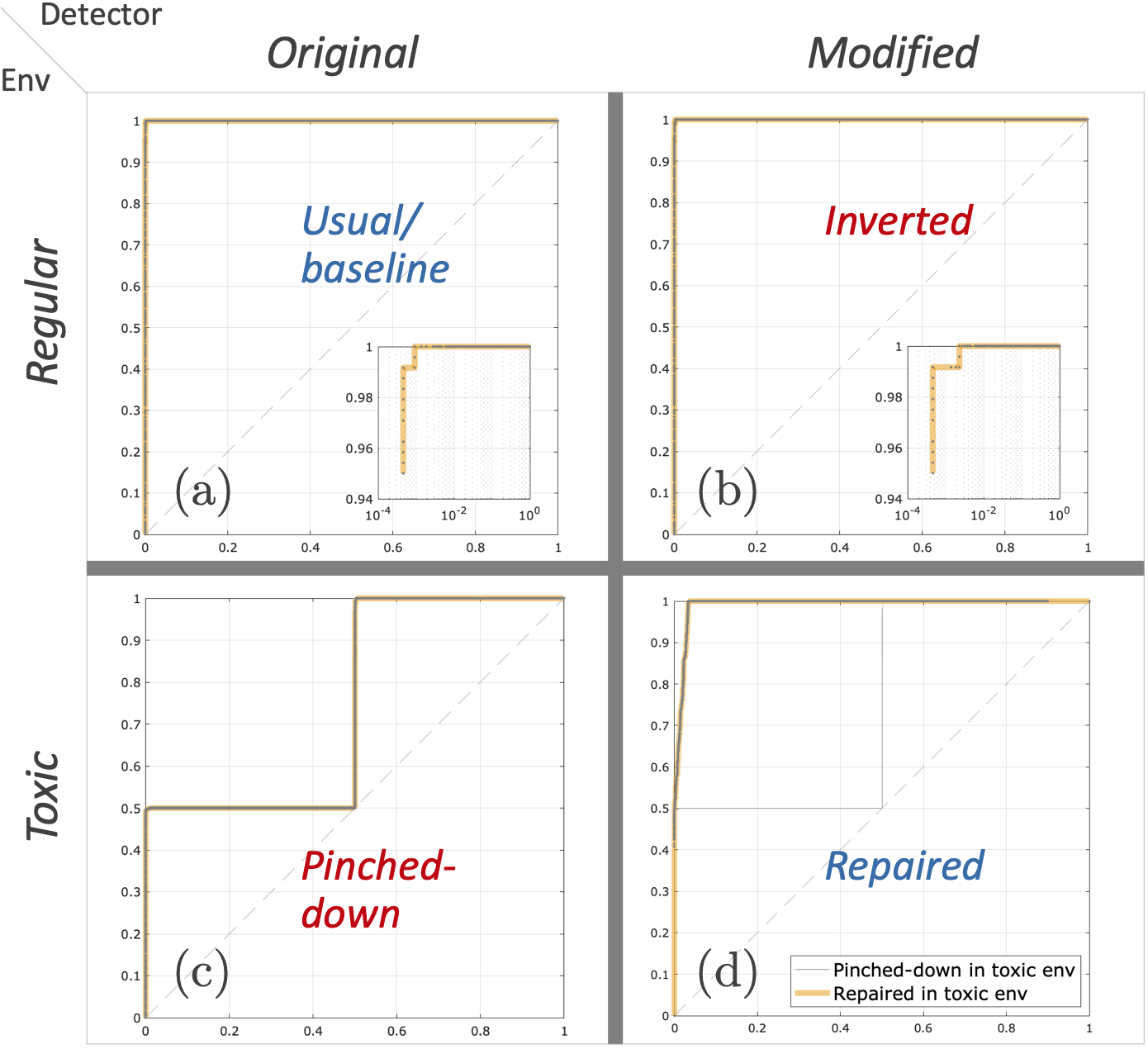}
\caption{Regular, inverted, pinched-down \& repaired ROCs in adversarial campaign against MNIST detector.}
\label{fig:ROC_summary}
\end{figure}

\begin{figure}
\centering
\includegraphics[width=3.38in]{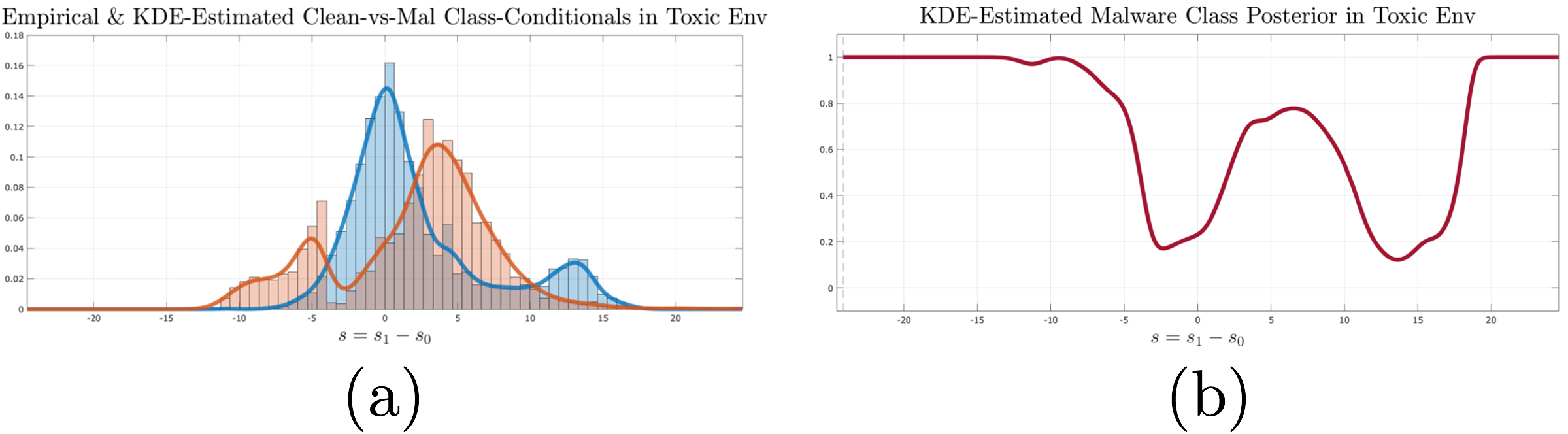}
\caption{Empirical \& smoothed class conditionals and equiprobable posterior for use in repair of the malware detector under special adversarial campaign.}
\label{fig:mal_detector_KDE_smoothed}
\end{figure}

\begin{figure}
\centering
\includegraphics[width=3.38in]{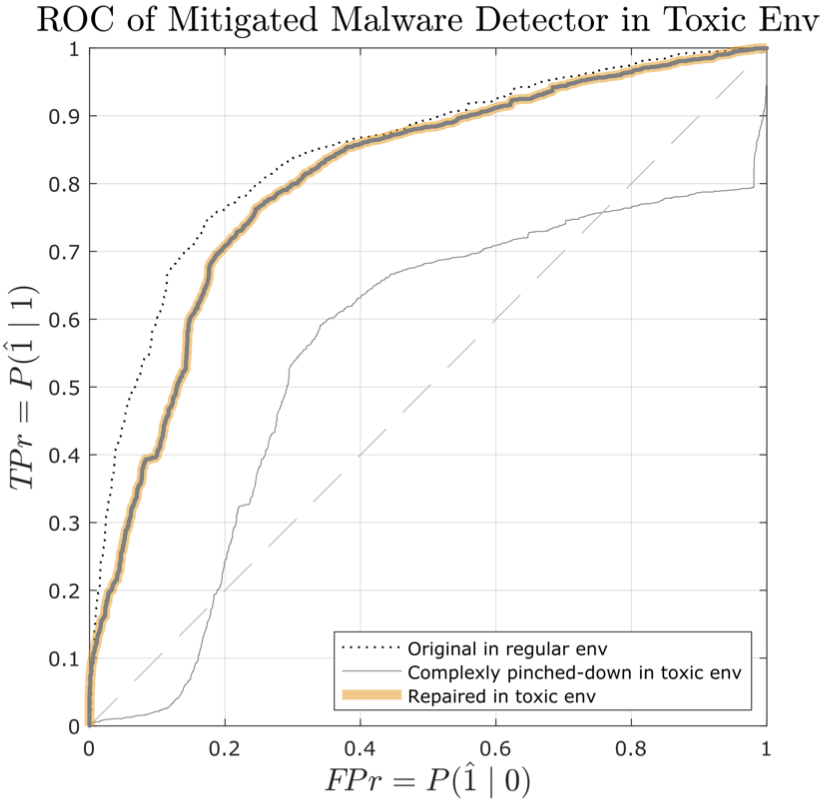}
\caption{Regular, pinched-down \& repaired ROCs in special adversarial campaign against malware detector.}
\label{fig:ROC_malware_detector}
\end{figure}

Aided by 2160 of the adversarially discovered FPs and 240 of the FNs to mimic the natural 9:1 class lopsidedness in the regular environment, we estimated decision-reversal interval as [--1.6,1.6] (graphically from intersection of empirical conditional histograms, much like green vs purple curves in Figure 2(e)), and generated the empirical ROC over the 2400 regular samples. Figure 9(b) inset confirms that the resulting ROC is inverted. It appears to be small harm but there is actually almost an order-of-magnitude larger FP rate; this difference is critical in the field.

The two class-conditional PDFs were estimated using Gaussian kernel and bandwidth $B$. This is equivalent to fitting a Gaussian mixture distribution where means equal all individual data points, and covariances equal the shared constant $B^2$. During a very toxic adversarial campaign where half of all samples become turbid, KDE-smoothed PDFs from aggregated data at the desired e-vs-d proportions reveal 3 crossovers (Figure 8(a)). The corresponding equiprobable posterior crosses the 0.5 threshold 3 times (Figure 8(b)). Operating the original unmitigated detector yields the pinched-down ROC (Figure 9(c)). In contrast, passing the scores thru the mitigated posterior function yields the repaired ROC (Figure 9(d)). Figure 9 confirms the inversion, pinch-down, and repair as predicted by the theory. We note however that the empirical nature of the construction makes ROCs look like staircases and it's impossible to discern convexity of the inversion from concavity of the pinch-down (something we know only from the theory).

\subsection{Malware Detection from Raw Bytes}

Hand-crafted feature engineering for static malware detection takes substantial expertise and years to develop. Increasingly, deep learning alternatives are showing promise as end-to-end feature learners-plus-classifiers, trained from raw binary file examples \cite{Raff2017, krcal2018}. We now verify the adversarial campaign health management framework using a pre-production model intended for an IIoT ``ICSP Neural'' USB scanning device. The model was made purposely suboptimal (with regular ROC curve far from upper-left corner) in order to better observe the manifestations of the theory. It contains a deep convolutional neural network with \{embedding, 4 convolutional, 3 dense, softmax\} layers summarized in Appendix A4, trained on half a million raw executable files (originally aggregated from a mix of clean and malicious customer submissions and vendor feeds), XORed with a common byte for inoculation at rest, and spanning at least 6 months of age to encourage learning `invariant' features. This type of network is fed integers in [0,255] representing bytes of a file zero-padded or cropped to length 700,000 (as if it were a wide image that is only 1 pixel tall). The test dataset consisted of 2000 clean and 2000 malicious files sampled from a time split spanning one month after the model's training date.

One of the simplest adversarial attacks for binaries to circumvent malware protection is to append a crafted payload at the end of the file \cite{trustcom18dl}. These methods can append a binary string, backdoor legitimate files by adding a new section to the executable (either as data or code), or use the resource part of the file when modifying already compiled code. Many far more sophisticated attacks are available \cite{anderson2018learning, suciu2019}.

In the present focus of research, the quality of the attack is less important than just finding misclassification-inducing perturbations, so we used the brute-force algorithm in Appendix A3 that appends fixed or random chunks until the model flips its decision (to within a 1000-trial count tolerance). A \textit{high-confidence} campaign was defined as a set of new binaries bypassing the model with pseudo-probability output above 0.97. Drawing seeds from the size-4000 test set, 512 adversarial FPs and 524 adversarial FNs were created this way.

Figure~\ref{fig:mal_detector_KDE_smoothed} shows the class-conditional likelihoods in the toxic environment, which are characterized (when pegged to the regular minimum balanced error score threshold 2.2) by 357 unforced + 512 forced FPs, plus 498 unforced + 524 forced FNs, totaling 1891 errors and thus a 62.5\%-37.5\% clear-to-turbid ratio. The clean (class 0) conditional is estimated from

\begin{equation}
{p_{{\rm{toxic}}}}(s|0) \approx {\textstyle{1 \over {{n_0}}}}\sum\limits_{i = 1}^{{n_0}} {{\textstyle{1 \over {{B_0}}}}} K\left( {{\textstyle{{s - s_i^{(0)}} \over {{B_0}}}}} \right)\:,
\end{equation}

where $K(\cdot)$ is the standard Gaussian kernel, $s_i^{(0)}$ are the unfolded decision scores ($2^\text{nd}$-unit softmax preactivation minus $1^\text{st}$) under class 0, and $B_0 = 0.687$ is the bandwidth from Silverman's estimate. The malware (class 1) conditional was similarly obtained with $B_1 = 0.928$. Unlike in previous situations, this special high-confidence campaign has adversarial scores dominating at one tail of each distribution, with no adversarial scores in the interval [--4,4]. That creates a complex posterior with 4 crossovers with respect to 0.5, to be used for repair (Eq. (15), Figure ~\ref{fig:mal_detector_KDE_smoothed}(b)).

Figure~\ref{fig:ROC_malware_detector} shows the devastating effect of this campaign on ROC and how much could be mitigated. Instead of a single pinch-down somewhere along the midsection of the curve, a composite of pinch-down and inversion brings the whole curve down around the chance line. Passing the original decision scores through the KDE-formed posterior brings the whole curve back to at least a gracefully degraded state.

We have seen that the health of both image-recognition and malware-detection components of industrial systems could be managed using our ROC-centric methodology, but this requires some ``server side,'' even if lagged, for label estimates. A subtle implication is that the introspective ``client-side'' monitoring alternative in Section 3 (where the device itself could declare adversarial campaign if too many decision scores are landing in an uncertain band) wouldn't work with the high-confidence adversarial campaign here. Adversarial actors aren't required to play by the small delta-perturbation rule as much in security as it is with natural images \cite{gilmer2018motivating}. Semantic proximity between a regular image $\textbf{\textit{x}}$ and adversarial counterpart $\textbf{\textit{x}}'$ means that humans wouldn't perceive them as belonging to different classes, thus perturbations tend to be small, placing $\textbf{\textit{x}}'$ near model's uncertain boundaries. For malware, semantic proximity only means that $\textbf{\textit{x}}'$ will still behave maliciously (or clean will stay clean), not so much that is has to closely resemble the input $\textbf{\textit{x}}$. This is manifested as adversarial distribution modes that are central in Figure~\ref{fig:KDE_smoothed} vs at the extreme ends in Figure~\ref{fig:mal_detector_KDE_smoothed}. Knowledge of this asymmetry can help guide simulations for preemptive domain adaptation.

\section{Conclusions}

The common misunderstanding surrounding what to do about adversarial inputs that fool detectors can be cleared by fixing the ``regular-vs-adversarial'' dichotomy and by recognizing the difference between one-off/per-trial basis protection vs adversarial campaign mitigation. Our investigation suggests that universal pre-hardening defenses are impossible without paying a price in accuracy of the original model operating in its regular environment.

We introduced turbidity detection, campaign mitigation, and preemptive domain adaptation as conceptual frameworks leading to practicable detector health management solutions. The theory yielded previously unreported results about ROC inversion, pinch-down, and repair in the context of adversarial threats to deep neural networks increasingly used in industry. Though not tested here, results should generalize to non-neural detectors such as ensembles of decision trees, as long as there is access to an internal score.

It should be understood that our method is not a panacea to shield or empower a model; what it does is optimally mitigate the damage (dramatically so for some ROC operating points) caused by adversarial toxicity that the original model wasn't designed to tackle on its own.

% Bibliography
\bibliographystyle{apacite}
\bibliography{ijphm19_paper_rev.bib}

\section*{Biographies}

% Tricky formats by trial-n-error for bios with picture
\begin{figure}[h!]
\begin{wrapfigure}[6]{l}[0pt]{0.7in}
\vspace{-15pt}
\centering
\includegraphics[width=0.83in,height=0.83in,clip,keepaspectratio]{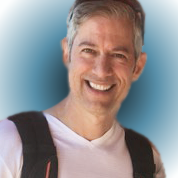}
\end{wrapfigure}
\noindent\textbf{Javier Echauz} is a Sr Machine Learning Research Scientist \& Technical Director at Symantec Corporation. He obtained a PhD at the interface of electrical, computer, and biomedical engineering from Georgia Tech.
\end{figure}

\begin{figure}[h!]
\begin{wrapfigure}[6]{l}[0pt]{0.7in}
\vspace{-20pt}
\centering
\includegraphics[width=0.83in,height=0.83in,clip,keepaspectratio]{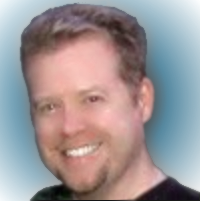}
\end{wrapfigure}
\noindent\textbf{Keith Kenemer} is a Principal Research Engineer at Symantec Corporation. He obtained an MSEE degree at the Georgia Institute of Technology.
\end{figure}

\begin{figure}[h!]
\begin{wrapfigure}[6]{l}[0pt]{0.7in}
\vspace{-15pt}
\centering
\includegraphics[width=0.83in,height=1in,clip,keepaspectratio]{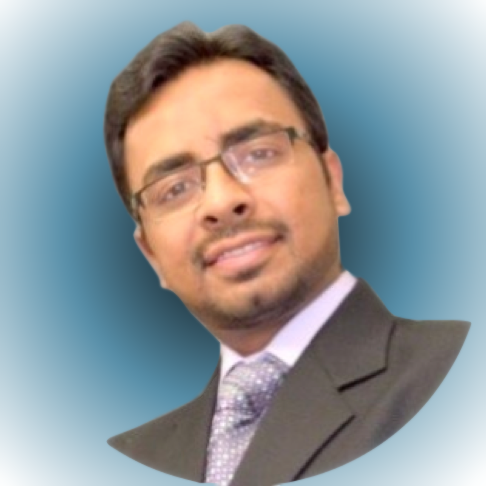}
\end{wrapfigure}
\noindent\textbf{Sarfaraz Hussein} is a Research Scientist and Engineer at Symantec Corporation. He obtained a PhD in Computer Vision /Medical Imaging at the University of Central Florida.
\end{figure}

\begin{figure}[h!]
\begin{wrapfigure}[6]{l}[0pt]{0.7in}
\vspace{-18pt}
\centering
\includegraphics[width=0.83in,height=1in,clip,keepaspectratio]{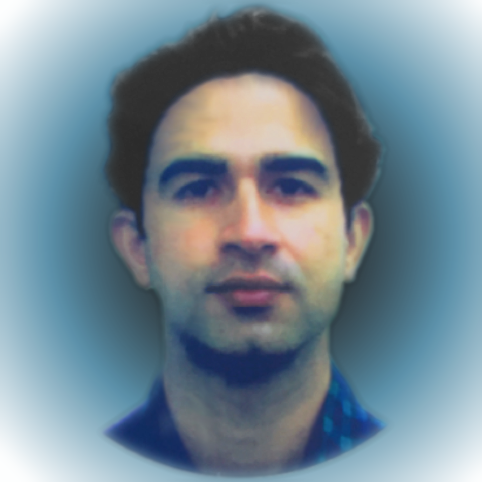}
\end{wrapfigure}
\noindent\textbf{Jay Dhaliwal} is a Sr Machine Learning Research Engineer at Symantec Corporation. He obtained an MS in Computer Science from University of Massachusetts Amherst.
\end{figure}

\begin{figure}[h!]\end{figure}
\begin{figure}[h!]
\begin{wrapfigure}[6]{l}[0pt]{0.7in}
\vspace{-18pt}
\centering
\includegraphics[width=0.83in,height=1in,clip,keepaspectratio]{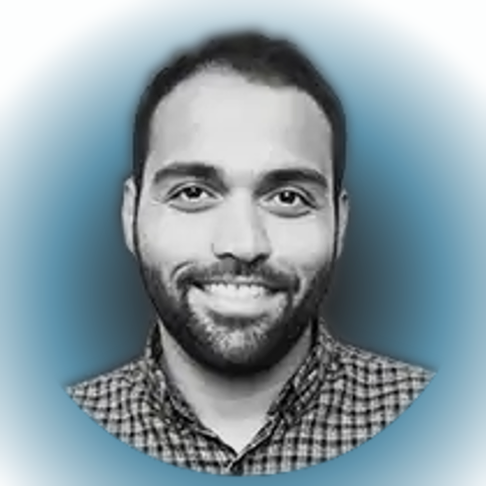}
\end{wrapfigure}
\noindent\textbf{Saurabh Shintre} is a Sr Principal Research Engineer at Symantec Corporation. He obtained a PhD in Electrical \& Computer Engineering from Carnegie Mellon University.
\end{figure}

\begin{figure}[h!]\end{figure}
\begin{figure}[h!]
\begin{wrapfigure}[6]{l}[0pt]{0.7in}
\vspace{-18pt}
\centering
\includegraphics[width=0.83in,height=1in,clip,keepaspectratio]{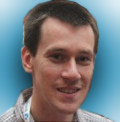}
\end{wrapfigure}
\noindent\textbf{Slawomir Grzonkowski} is a Sr Principal Research Engineer at Symantec Corporation. He obtained a PhD in Computer Science from the National University of Ireland.
\end{figure}

\begin{figure}[h!]\end{figure}
\begin{figure}[h!]
\begin{wrapfigure}[6]{l}[0pt]{0.7in}
\vspace{-23pt}
\centering
\includegraphics[width=0.83in,height=1in,clip,keepaspectratio]{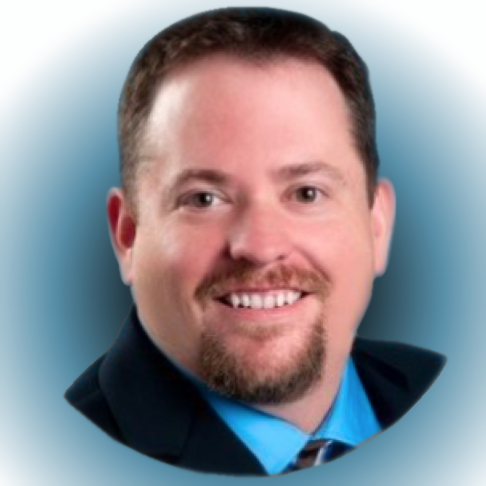}
\end{wrapfigure}
\noindent\textbf{Andrew Gardner} is a Senior Technical Director at Symantec Corporation. He obtained a PhD in Electrical \& Computer Engineering from Georgia Tech.
\end{figure}

\begin{figure}[h!]\end{figure}

\section*{Appendix}

\subsection*{A.1 Exact ROC for Nonmonotonic Posterior}
Pseudocode for generating the exact multi-branched ROC curve without data, given possibly nonmonotonic class-t posterior $P(\text{t}|s)$, and $F_{\neg \text{t}}$ (non-target) vs $F_\text{t}$ (target) CDFs.

\begin{algorithmic}
\Require pre, a vector of preimages of $P(\text{t}|s)$ extrema (endpoints of segments to search for roots)
\Ensure fpr, tpr arrays
\\
\State $N_{\text{seg}} \gets \text{len(pre)}-1$ \Comment{Number of segments}
\State $B \gets (1-(-1)^{N_{\text{seg}}})/2$ \Comment{0 or 1 per even or odd}
\State fpr, tpr $\gets [B ... B]$ \Comment{Broadcast initialization}
\Repeat
\State Set a threshold $\theta$ from grid or random sample
\For {seg in range(1,$N_{\text{seg}}+1$)}
    \State $c \gets \text{root}(P(\text{t}|s)-\theta)$ \Comment{Find root in this segment}
    \State $\text{fpr}_\theta \gets \text{fpr}_\theta + (-1)^{\text{seg}}F_{\neg \text{t}}(c)$
    \State $\text{tpr}_\theta \gets \text{tpr}_\theta + (-1)^{\text{seg}}F_\text{t}(c)$
\EndFor
\Until Enough $\theta$ coverage
\end{algorithmic}

\subsection*{A.2 Methods to Aid Replication}
The reader can quickly verify shapes of distributions and ROCs in this paper (even if empirically without the benefit of A.1) via Monte Carlo methods. The reference regular DGP in Section 2 can be functional-programmed (in MATLAB/ Mathematica/ R style) directly as
{\tiny\begin{verbatim}
  pd = makedist('Logistic','mu',0,'sigma',1)
  pdf0 = @(s)pd.cdf(-s).*unifpdf(s,-10,10)/.5
  pdf1 = @(s)pd.cdf(s).*unifpdf(s,-10,10)/.5
  cdf0 = @(s)integral(@(q)pdf0(q),-Inf,s)
  cdf1 = @(s)integral(@(q)pdf1(q),-Inf,s)
\end{verbatim}}
Then a functional plotter with adaptively sampled domain and parametric option will graph ROC directly, e.g., Figure 2(d) is {\small\texttt{fplot(@(s)cdf1(-s), @(s)1-cdf1(s))}}. An empirical version of this, e.g., in Python\footnote[1]{As of this writing, there is a deprecated 1D-only plotter in {\small\texttt{scipy}}, and a {\small\texttt{sympy}} approach that is limited to its known set of functions.}, can generate $n$ randomly sampled scores emitted by the DGP. A 1-D dataset \textbf{X},\textbf{y} consists of matrix \textbf{X} being a length-$n$ array with corresponding labels \textbf{y} = sign(\textbf{X}+\boldmath${\xi}$\unboldmath), where the noise array is sampled from $Logistic(0,1)$. Now index into data to obtain the corresponding class-conditional histograms of \textbf{X}[\textbf{y}==–1] vs \textbf{X}[\textbf{y}==1] (Figure 2(a)), and an empirical ROC from {\small\texttt{roc\_curve}}(\textbf{y},\textbf{X}).

The same regular dataset \textbf{X},\textbf{y} can be indexed to obtain turbidity e-vs-d conditionals, with histograms of
\[\begin{array}{l}
\textbf{X}_\text{e} = \textbf{X}[ (\textbf{y}==–1\:\&\:\textbf{X}<0) \mid (\textbf{y}==1\:\&\:\textbf{X}>=0) ]\\
\textbf{X}_\text{d} = \textbf{X}[ (\textbf{y}==–1\:\&\:\textbf{X}>=0) \mid (\textbf{y}==1\:\&\:\textbf{X}<0) ]
\end{array}\]
(Figure 2(e)). In the regular environment, this will yield approximately $F_0(0)n$ samples (e.g., 9303 when $n$=10,000) of `clear' vs only $F_1(0)n$ samples (e.g., 693) of `turbid'. In adversarially altered DGP environments, aspects like accuracy calculation and the marginal histogram (Figure 2(f)) seen by the model can be simulated by rebalancing the data via over- and/or under-sampling by a rational factor that approximates $F_0(0)/F_1(0)$ = 13.4., e.g., oversampling the minority class by 13 or 14. The corresponding \textbf{y}$_\text{e}$ and \textbf{y}$_\text{d}$ maliciousness labels can be used to unit-test/verify that accuracy $\approx 50\%$ (1$^\text{st}$ group all correct, 2$^\text{nd}$ all wrong).

Regarding turbidity detection, to avoid confusion with \textbf{y} maliciousness labels --1s vs +1s (or 0s vs 1s), we could assign 2s to the `clear' class and 3s to the `turbid' class. Then the e-vs-d ROC (Figure 2(h)) can be empirically obtained from true labels [2, 2, \dots($n_\text{e}$ times), 3, 3, \dots($n_\text{d}$ times)], predicted soft labels $P(\text{d} \mid [\textbf{X}_\text{e}, \textbf{X}_\text{d}])$ from Eq. (10), and `3' as the target class in the function {\small\texttt{roc\_curve}}.

The ROC pinch-down (Figure 4(b)) can be empirically verified by sending preactivation scores to the function {\small\texttt{roc\_curve}} when the DGP is adversarially toxic, e.g., [\textbf{X}$_\text{e}$, repeat(\textbf{X}$_\text{d}$ 14 times)]. To verify ROC inversion (Figure 3(b)) or repair (Figure 6(b)), the scores are first passed thru a possibly nonmonotonic posterior function before sending to {\small\texttt{roc\_curve}}. For inversion, the posterior is $P_\text{aug}$ in Section 2.3 under a regular DGP. For repair, the posterior is Eq. (15) under an adversarially toxic DGP.

\subsection*{A.3 High-Confidence Adversarial Attack}
A brute-force append attack to generate high-confidence adversarial FPs or FNs from raw binaries.
\begin{algorithmic}
\Require file, max\_size, model
\Ensure file\_new
\\
\Function{Check}{X}
    \State pred $\gets$ model.predict(X)
    \If {pred $>$ pred\_max}
        \State file\_new $\gets$ X
        \State pred\_max $\gets$ pred
    \EndIf
    \If {pred\_max $>$ 0.97} Signal break
    \EndIf
\EndFunction
\\
\State file\_new $\gets$ file + (max\_size -- file.size) * [0x00]
\State pred\_max $\gets$ model.predict(file\_new)
\For {i in range 0x01 to 0xff} \Comment{Phase 1: Fill constant}
    \State X $\gets$ file + (max\_size -- file.size) * [i]
    \State \Call{Check}{X}
\EndFor
\If {pred $\le$ 0.97} \Comment{Phase 2: Random chunk if needed}
    \For {i in range(0,1000)}
        \State Pick random chunk C in file\_new(file.size:max\_size)
        \State Pick random vector V sized as C
        \State X $\gets$ Replace chunk with C+V 
        \State \Call{Check}{X}
    \EndFor
    Failed to find high-confidence example
\EndIf
\end{algorithmic}

\subsection*{A.4 Malware Detector Model Summary}
The deep neural network investigated was a sequential Keras-wrapped TensorFlow model with 840,882 parameters as summarized below.

%\verbatimfont{\normalfont} DOES change to roman
{\tiny\begin{verbatim}
___________________________________________________________
Layer (type)              Output Shape           Param #  
===========================================================
input (InputLayer)        (None, 700000, 1)      0        
___________________________________________________________
reshape_1 (Reshape)       (None, 700000)         0        
___________________________________________________________
embedding (Embedding)     (None, 700000, 8)      2048     
___________________________________________________________
conv1 (Conv1D)            (None, 175000, 48)     12336    
___________________________________________________________
relu1 (Activation)        (None, 175000, 48)     0        
___________________________________________________________
conv2 (Conv1D)            (None, 43750, 96)      147552   
___________________________________________________________
relu2 (Activation)        (None, 43750, 96)      0         
___________________________________________________________
temporal_max_pooling (MaxP(None, 10938, 96)      0        
___________________________________________________________
conv3 (Conv1D)            (None, 1368, 128)      196736   
___________________________________________________________
relu3 (Activation)        (None, 1368, 128)      0        
___________________________________________________________
conv4 (Conv1D)            (None, 171, 192)       393408   
___________________________________________________________
relu4 (Activation)        (None, 171, 192)       0        
___________________________________________________________
global_temporal_avg_poolin(None, 1, 192)         0        
___________________________________________________________
flatten (Flatten)         (None, 192)            0        
___________________________________________________________
fc1 (Dense)               (None, 192)            37056    
___________________________________________________________
selu1 (Activation)        (None, 192)            0        
___________________________________________________________
fc2 (Dense)               (None, 160)            30880    
___________________________________________________________
selu2 (Activation)        (None, 160)            0        
___________________________________________________________
fc3 (Dense)               (None, 128)            20608    
___________________________________________________________
selu3 (Activation)        (None, 128)            0        
___________________________________________________________
logits (Dense)            (None, 2)              258      
___________________________________________________________
output (Activation)       (None, 2)              0        
===========================================================
\end{verbatim}}

\end{document}